\documentclass{article}

\usepackage{PRIMEarxiv}

\usepackage[utf8]{inputenc} % allow utf-8 input
\usepackage[T1]{fontenc}    % use 8-bit T1 fonts
\usepackage{hyperref}       % hyperlinks
\usepackage{url}            % simple URL typesetting
\usepackage{booktabs}       % professional-quality tables
\usepackage{amsfonts}       % blackboard math symbols
\usepackage{nicefrac}       % compact symbols for 1/2, etc.
\usepackage{microtype}      % microtypography
\usepackage{lipsum}
\usepackage{fancyhdr}       % header
\usepackage{graphicx}       % graphics
\graphicspath{{media/}}     % organize your images and other figures under media/ folder

\usepackage{amsmath,graphicx,subcaption,multirow}
\usepackage[labelformat=simple]{subcaption}

\DeclareCaptionLabelFormat{subcaptionlabel}{\normalfont(\textbf{#2}\normalfont)}
\title{Bidirectional Representations for Low-Resource Spoken Language Understanding
\thanks{\textit{\underline{Citation}}: 
\textbf{Q. Meeus, M.-F. Moens, H. Van hamme. Bidirectional Representations for Low-Resource Spoken Language Understanding. Appl. Sci. 2023, 13(20), 11291; DOI:10.3390/app132011291}} 
}

\author{
  Quentin Meeus \\
  Dept. Computer Science \\
  Dept. Electrical Engineering \\
  KU Leuven
  Belgium\\
   \And
  Marie Francine Moens \\
  Dept. Computer Science \\
  KU Leuven \\
  Belgium \\
   \And
  Hugo Van hamme \\
  Dept. Electrical Engineering \\
  KU Leuven \\
  Belgium \\
}

\begin{document}
\maketitle

\begin{abstract}
Speech representation models lack the ability to efficiently store semantic information and require fine tuning to deliver decent performance. In this research, we introduce a transformer encoder--decoder framework with a multiobjective training strategy, incorporating connectionist temporal classification (CTC) and masked language modeling (MLM) objectives. This approach enables the model to learn contextual bidirectional representations. We evaluate the representations in a challenging low-resource scenario, where training data is limited, necessitating expressive speech embeddings to compensate for the scarcity of examples. Notably, we demonstrate that our model's initial embeddings outperform comparable models on multiple datasets before fine tuning. Fine tuning the top layers of the representation model further enhances performance, particularly on the Fluent Speech Command dataset, even under low-resource conditions. Additionally, we introduce the concept of class attention as an efficient module for spoken language understanding, characterized by its speed and minimal parameter requirements. Class attention not only aids in explaining model predictions but also enhances our understanding of the underlying decision-making processes. Our experiments cover both English and Dutch languages, offering a comprehensive evaluation of our proposed approach.
\end{abstract}
%%%%%%%%%%%%%%%%%%%%%%%%%%%%%%%%%%%%%%%%%%

\section{Introduction}
\label{sec:intro}
Commercial spoken language understanding (SLU) systems rely on cascading automatic speech recognition (ASR) with natural language understanding (NLU). We identify three main problems of such interfaces: (1) the ASR  errors are cascaded to the NLU system; (2) discrepancies arise between training and inference, as the NLU model is often not trained on spoken language, which differs in terms of phrasing but also includes acoustic elements such as stress and pitch that are lost after transcription; and (3) the resulting systems can be large, as both recent ASR and NLU models contain an increasing number of parameters. However, it is undeniable that there is a strong need for lightweight, robust personal assistants able to understand the nuances and intricacies of spoken language. Those systems should also work out of the box, without requiring much effort from the user.

Since their introduction in 2017, transformers \cite{Vaswani2017} have taken over the field of natural language processing (NLP). The key to their success is their ability to process long sequences, attend to relevant pieces of information present in the input and learn qualitative representations that can be used for other purposes. This last aspect is especially true for masked language models (MLMs), where a model is trained to reconstruct a partial input in a ``fill in the gap'' fashion. In this setting, words are masked from the input sentence, and the model must predict what was removed. While doing so, transformers learn syntactic and semantic information, such as parts of speech and language composition \cite{jawahar-etal-2019-bert}. Unlike conditional language models (CLMs), which only have access to past information to generate a prediction, MLMs can benefit from left and right contexts \cite{Devlin2019,BECTRA}. Furthermore, whereas CLMs are limited to the generation of one token at a time conditioned on the previously generated tokens, MLMs can fill multiple masked tokens in parallel. Nonetheless, CLMs are still widely popular for language generation tasks.

In speech processing, transformers have also been widely adopted by the research community \cite{Karita2019Compare}. A major difference with text-based transformers is the addition of a convolutional front end to process continuous sequences \cite{Dong2018,Mohamed2020}. Its role is to learn local relationships and encode positional information. Both encoder-only and encoder--decoder transformers have been experimented with. Large self-supervised encoder-only models have shown impressive performance in many speech processing tasks \cite{wav2vec2,hubert}. However, the amount of semantic information contained in these encoders is limited or so deeply entangled that extracting it is challenging, and fine tuning the encoder is necessary to achieve satisfactory performance \cite{WangBoumadane2021}. In contrast, encoder--decoder models split the representation task over two distinct modules: the encoder models the acoustic information contained in speech, while the decoder learns more language related features such as syntax and semantics \cite{espnet-slu}. The encoder--decoder relies only on attention to align speech and text. \cite{hybridctcatt} proposed the integration of connectionist temporal classification (CTC) \cite{Graves2006}, a technique that has been well researched in speech recognition. The CTC objective in the hybrid transformer enforces a monotonic alignment between speech and text, resulting in improved robustness and faster convergence \cite{hybridctcatt}. Another notable contribution is that of \cite{BECTRA}, who concatenated the outputs of a speech encoder and BERT, which was then passed to a prediction network. They also noted the benefit of using left and right context, as opposed to models learning only from past values. Finally, attempts have been made to augment large language models with speech capabilities, although more research is needed to achieve competitive performance \cite{gao2022}. Despite considerable improvement in speech modeling in recent years, speech representation models do not show the same ability as text-based language models to efficiently store semantic information, and a considerable amount of fine tuning is necessary to achieve decent performance \cite{WangBoumadane2021}.

In this research, we explore representation models of speech and demonstrate their usefulness for SLU, with a particular focus on low-resource solutions, when only a handful of examples is available to train and validate the models. We propose an architecture for pretraining of bidirectional speech transformers on a surrogate ASR task with the goal of learning qualitative representations that prove useful for solving language understanding tasks. We evaluate the resulting embeddings in an SLU task, intent recognition, where the objective is to, given a spoken command, identify the intent and any argument necessary. For example, the sentence ``switch the lights in the kitchen to blue'' would result in an intent of ``lights'' and arguments of ``room = kitchen'' and ``color = blue'', while ``play Beyonce'' has an intent of ``music'' and an argument of ``artist = Beyonce''. For completeness and comparison purposes, we also include results obtained after partially fine tuning the model.

We are particularly interested in low-resource scenarios, where we control the number of examples used for training and validation to study the limits of what can be learned from the representations. When information is stored efficiently, few examples are necessary to learn how to extract it. In contrast, noisy embeddings require many examples to extract relevant information. For the SLU module, we propose class attention as a drop-in replacement of LSTMs. Class attention assigns a score to each position that relates to the importance of the corresponding token to predict the intent and arguments. Aside from the advantages of its small footprint, it provides us with a method to understand why a model makes predictions. In the remainder of this article, we first detail the architecture of the different components of the model. Then, we lay out the experimental setup and methodology. Finally, we evaluate the models and analyze the representations in detail.

%%%%%%%%%%%%%%%%%%%%%%%%%%%%%%%%%%%%%%%%%%
\section{Methods}
\label{sec:model}

\begin{figure}[h]
\includegraphics[width=\linewidth]{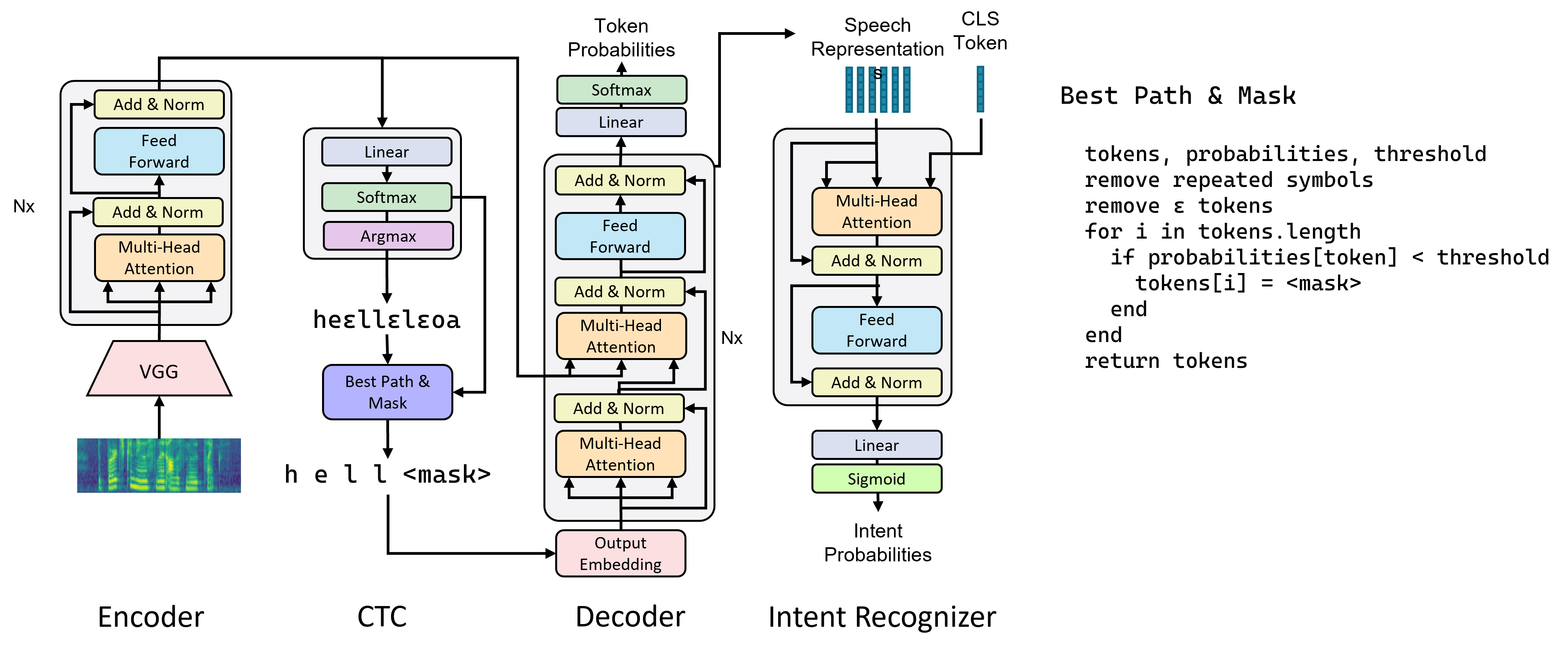}
\caption{The encoder processes speech features. The CTC module makes a rough prediction of the text, where low-probability tokens are masked. The decoder attends to the masked transcription and to the encoder's output and predicts missing symbols. The generated representations are used as input to the downstream model to predict the intent.}
\label{fig:architecture}
\end{figure}

Building on the work of \cite{MaskCTC2020}, we propose an encoder--decoder architecture with a multiobjective training strategy to learn bidirectional representations of speech (Figure \ref{fig:architecture}). We evaluate the features learned in a downstream SLU task: intent prediction. The encoder, as presented in Section \ref{sec:model:encoder}, learns acoustic representations of speech. The sequences of acoustic unit representations are used in two modules in parallel: they are mapped to output symbols with a classification layer and optimized with CTC (Section \ref{sec:model:ctc}), and they are processed by a bidirectional transformer decoder to learn linguistic features (Section \ref{sec:model:decoder}). These features learned by the decoder are processed by an intent recognizer to obtain the final output (Section~\ref{sec:model:downstream}).

\subsection{Encoder}
\label{sec:model:encoder}
The encoder processes Mel-scaled filter banks and transforms them into a sequence of acoustic embeddings ($h_{\text{enc}}$). The module is composed of a VGG-like convolutional front end followed by a transformer encoder. The role of the CNN is twofold: decrease the frequency and time dimensions and learn the local relationships in the sequence of speech features. This replaces the positional encoding traditionally required by transformers to keep track of the sequence order \cite{Mohamed2020}. The CNN front end compresses a sequence of speech features of length $T$ to a smaller sequence of length $T' \approx T / 4$. The transformer encoder is composed of multiple blocks of multihead self-attentions and position-wise fully connected feedforward networks. Each block has a residual connection around both operations, followed by layer normalization. In this setting, the encoder cannot be trained in a standalone process.

\subsection{CTC} 
\label{sec:model:ctc}
Connectionist temporal classification \cite{Graves2006} is a method for optimizing sequence-to-sequence models. For each unit in the input sequence, CTC predicts a probability distribution over the vocabulary, consisting of the set of output symbols and a blank token. The predictions are assumed to be conditionally independent. A simple algorithm allows a sequence of CTC tokens to be reduced to a sentence by first removing repeated symbols, then removing blank tokens (Figure \ref{fig:architecture}). The CTC loss is computed by summing the negative log likelihood of all the alignments that result in the correct sentence. For decoding, we opt for a greedy algorithm, which takes the token predicted with the highest probability at each time step. Although more advanced techniques such as beam search or weighted finite-state transducers \cite{ctcdecode} can improve the CTC prediction considerably, \cite{ctcdecode} found that the largest gains were observed in substitution errors, with substantially lower gains for insertions and deletions. For this reason, we prefer a fast decoding technique and let the decoder fix the substitution errors in the rough prediction.

\subsection{Decoder}
\label{sec:model:decoder}
Although an encoder equipped with CTC is perfectly able to transcribe speech, it does so by assigning a token to each position independently. \cite{hybridctcatt} reported that the addition of a decoder integrates language modeling in the learning process. Indeed, the decoder combines the acoustic cues of the encoder with the previous tokens to predict the next one. Here, we opt for a bidirectional transformer decoder instead of a left-to-right model. Left-to-right decoders predict one token at a time and stop after producing a special end-of-sequence token. This method is slower because a prediction depends on the previously predicted tokens. Additionally, the network only has access to previous tokens, thus limiting the amount of information available to make a prediction. In modern NLP, bidirectional architectures have become popular to learn powerful representations that make use of both left and right contexts \cite{Devlin2019,ELMo}. For instance, BERT \cite{Devlin2019} is optimized with a masked language modeling objective by randomly masking some tokens in the input sequence. In ASR, a similar approach was adopted by \cite{MaskCTC2020}, who obtained a rough text prediction with CTC, which was refined by masking low-probability tokens and letting the decoder predict the missing items in the sequence, similarly to \cite{MaskPredict}. One issue with these models is the decoder's initial state definition. As the output sequence is unknown at the inference time, it is necessary to provide the decoder with a template consisting of a sequence of mask tokens of the same length as the target sequence. The decoder is not able to add or remove symbols to adjust the sequence length, and the correct length is unknown at the inference time. \cite{MaskPredict} presented a solution that involves predicting the length as part of the prediction task of the encoder. Then, during inference, the decoder predicts multiple candidates of different lengths corresponding to the predicted lengths with the highest probability. \cite{MaskCTC2020} assumed the greedy CTC prediction to be of correct length, although they showed that it leads to weaker transcriptions. In this work, we focus on learning rich representations for spoken language understanding. We thus work around this issue, as we assume that the length is not essential for our purpose. We follow \cite{MaskCTC2020} for the architecture of the decoder. We use the output of the penultimate layer as the linguistic representations of speech.

\subsection{Intent Recognizer}
\label{sec:model:downstream}
In the SLU datasets, intent is represented as a certain action (e.g., move, grab, turn, etc.) to which are attached a number of arguments (forward, fast, left, etc.). We encode the full intent string as a multihot vector, where each bit corresponds to either one of the possible actions or an argument value. The representations obtained from the decoder are summarized with a stack of class attention layers \cite{ClassAtt}. Class attention combines a sequence of representations into a class embedding ($x_{\text{CLS}}$), that is, a learned vector of the same dimensions as the representations. During training, the module learns to identify patterns in the input features that are predictive of the class labels. Similarly to other transformers, a class attention layer is composed of two sublayers: an attention layer and a point-wise feedforward layer. The input to the layer is normalized before processing to stabilize training, as prescribed by \cite{LayerNorm}. We define the queries in the original notation according to \cite{Vaswani2017} as the following class embedding: $Q = x_{\text{CLS}}$. The keys and values are linear projections of the elements in the input sequence of features ($x$): $K = W_k \cdot x + b_k$ and $V = W_v \cdot x + b_v$. In contrast to \cite{ClassAtt}, we do not include the CLS token in the keys and values. We compute the attention weights by applying the softmax function to the inner product of the keys and queries, scaled by the square root of the dimension of each attention head: $A = \sigma(Q \cdot K^T / \sqrt{d/h})$, where $d$ is the total dimension of $x_{\text{CLS}}$, and $h$ is the number of attention heads. The output corresponds to $y = W_oAV + b_o$. $W_k$, where $b_k$, $W_v$, $b_v$, $W_o$ and $b_o$ are learned parameters. We train the model by minimizing a binary cross-entropy loss. Since not all combinations are possible, we enforce the output structure by selecting the valid combination that minimizes the cross-entropy with the predicted vector. 

%%%%%%%%%%%%%%%%%%%%%%%%%%%%%%%%%%%%%%%%%%
\section{Experiments}
\label{sec:exp}

\subsection{Datasets}
\label{sec:exp:data}

%\begin{itemize}[label=,labelsep=0mm]
 Corpus Gesproken Nederlands \cite{CGN} is a collection of recordings in Dutch and Flemish collected from various sources, such as readings, lectures, news reports, conferences, telephone conversations, etc., totaling more than 900 h. They are divided into 15~components (a to o) based on their nature. After removing short and overlapping utterances, we divide each component into three subsets to serve as training, validation and test sets, respectively. We leave out three components (a, c and d) because the quality of the transcriptions differs considerably from that of the other components. The subsets from the remaining components, totaling 415 h of speech, are concatenated to form the training, validation and test sets. 

Librispeech \cite{Librispeech} contains about a thousand hours of read English speech derived from audiobooks. The authors provide official splits for training, validating and testing the models, and this dataset serves as a reference in ASR. We use all 960 h for training.

Grabo \cite{Renkens2018} is composed of spoken commands, mostly in Dutch (one speaker speaks in English) intended to control a robot. There are 36 commands that were repeated 15~times by each of the 11 speakers. More precisely, the robot can perform eight actions (i.e., approach, move (relative), move (absolute), turn (relative), turn (absolute), grab, point and lift), of which is further defined by some attributes (e.g., the robot can move forward or backward and rapidly or slowly). We follow the methodology applied in \cite{Renkens2018}: For each speaker, we create nine datasets of different sizes and divide each dataset in five~folds for cross validation. The target is represented as a binary vector of 31 dimensions.

Patience \cite{Patcor} is derived from a card game in which a player provides vocal instructions to move cards between different tiles on the table. The dataset contains recordings from eight different Flemish speakers. The intent of the player is represented as a binary vector. Again, we use the same splitting methodology \mbox{as \cite{Renkens2018}}, keeping only the 31 most represented classes. It should be noted that without visual input, this task is quite difficult, even for a human operator.

Fluent Speech Commands \cite{Fluent} contains 30,043 utterances from 97 English speakers. Each utterance is a command to control smart home appliances or a virtual assistant. The challenge splits that we use in this work were proposed by \cite{MASE}. Two test sets are provided, in which specific speakers or utterances are separated from the training and validation sets. We also explore data shortage scenarios, where only a portion of the training set is available.

 SmartLights \cite{Smartlights} is composed of 1660 commands to control smart lights uttered by speakers with various accents and of various origins. Again, we use the challenge split proposed by \cite{MASE}.
%\end{itemize}

\subsection{Pretraining}
\label{sec:exp:pt}
We pretrain our encoder--decoder model on an ASR task with a hybrid objective that is the weighted sum of the CTC loss ($\L_{\text{ctc}}$) and the label-smoothing cross-entropy loss ($\L_{\text{dec}}$: $\L = \rho \; \L_{\text{ctc}} + (1-\rho) \; \L_{\text{dec}}${, where $\rho$ and $1-\rho$ are the weights of the CTC loss and the decoder loss in the total loss, respectively}). We use CGN (220 h) to pretrain the Dutch model and Librispeech (960 h) for the English model. Our objective is to train a representation model of speech that captures elements of language and semantics while remaining robust to irrelevant aspects such as speaker identity or background noise. Additionally, the information stored in the encodings should be readily accessible for processing by a downstream model. We choose ASR as a proxy task because of the availability of large datasets and because it is easier to generate rich representations containing elements of language than with self-supervised approaches, which typically require considerably more resources to achieve the same goal. However, the supervised objective is likely to discard information not directly useful for solving the task. Addressing this neglected information will be the focus of future work. During pretraining, we use the real tokens as input to be masked for the decoder. During inference, we use the CTC module to generate a rough prediction that is used as a template, and we mask out tokens predicted with a probability of less than 90\%. Similarly to MaskPredict \cite{MaskPredict}, we iteratively refine the input for a maximum of 10 steps. We pretrain the models on a GPU for 200 epochs, with batches of 32 examples, and accumulate the gradients over 8 iterations, which provides an effective batch size of 256. We use the Noam learning rate scheduler \cite{Vaswani2017}. We experimentally set $\rho$ to $0.3$ by measuring the ASR accuracy on a held-out validation set.

\subsection{Training}
\label{sec:exp:train}
After pretraining the representation model, we freeze all the layers to train the spoken language understanding module. The training objective is to minimize the binary cross entropy between the predictions and the multihot-encoded targets. Freezing the representation model gives us the opportunity to evaluate the predictive power of the representations. At this point, the pretrained encoder--decoder from described in Section \ref{sec:exp:pt} has not been exposed to any training example from the SLU datasets, nor does it know about the output structure of the underlying task (number of classes, etc.). We train the models with the Adam optimizer \cite{Kingma2017} for a maximum of 200 epochs, with early stopping. The batch size is 512. We set the learning rate to $0.005$, except for the Smartlights dataset, for which we found that a slower learning rate of $0.001$ achieved better results.

\subsection{Fine Tuning}
\label{sec:exp:ft}
Our main objective with this research is to learn representations that perform well without fine tuning. However, to provide a good basis for comparison with previous research and to quantify how much can be gained by updating the main model, we perform a few fine-tuning steps at a low learning rate while unfreezing some or all layers of the representation model's decoder. This operation can be seen as specializing the representation model for the specific aspects of the downstream task, often at the expense of generality. {In practice, we found that unfreezing the whole decoder was more tedious and sometimes led to overtraining. Unfreezing the last four layers generally yielded the best results.} As discussed in Section \ref{sec:model:decoder}, we use the CTC module to produce the initial sequence in which low-probability tokens are masked. We perform only one pass with the decoder to generate the representations used by the SLU module.

\subsection{Evaluation}
\label{sec:exp:eval}
{We choose specific metrics to assess the performance of our models, primarily focusing on accuracy and F1 score. Accuracy measures the proportion of correctly predicted instances, encompassing both intents and slots. The F1 score, which is computed as the harmonic mean of precision and recall, provides a more holistic evaluation. While we consider the F1 score to offer a more comprehensive assessment, we also examine accuracy to facilitate comparisons with established models.

In our investigation into the model's performance within a low-resource setting, we deliberately limit the number of available training examples. This allows us to assess its adaptability and effectiveness under conditions of data scarcity.

Furthermore, to gain deeper insights into the representations generated by our model, we conduct a qualitative analysis. We achieve this by visualizing the representations using t-SNE \cite{tsne}, a technique that helps identify clusters of examples, shedding light on the structure and organization of the learned features. We also visualize the attention weights to understand how the model makes certain predictions.}

\subsection{Hyperparameters}
\label{sec:exp:hp}
The transformer has 18 layers (12 encoder layers and 6 decoder layers) each with 4 attention heads, a hidden layer size of 256 and 2048 hidden units in the linear layer. We use dropout with a probability of 0.1 after each transformer layer. The front end has two 2D convolutional layers with kernel sizes of 3 $\times$ 3 and a stride of 2 $\times$ 2.
Then, the input dimension is divided by a factor of four along the time and frequency dimensions. The hyperparameters related to the architecture of the encoder--decoder were chosen according to \cite{MaskCTC2020}. The encoder--decoder model has 30.9 million parameters. The weights for scaling pf the different losses in the final loss function are chosen through experimentation using the pretraining data. We train our models for 200 epochs with a batch size of {256}. The models are trained with the Adam optimizer \cite{Kingma2017}. The learning rate linearly increases until reaching a maximum value of 0.4 after 25,000 steps, then decreases according to the Noam schedule \cite{Vaswani2017}. The downstream datasets were not used in any way to determine the value of the abovementioned hyperparameters.

The downstream models are composed of two class attention layers with four attention heads with 32 dimensions. The fully connected layer has 1024 units. The models are trained with batches of 512 examples for 100 epochs, with a learning rate equal to 0.005. The intent classification module has 890 thousand parameters.

\section{Results}
\label{sec:results}

We compare our models on basis of the accuracy score to remain consistent with previous research. {We select four baselines because they propose a similar approach as ours and present results on at least one of our selected SLU datasets.} End-to-end SLU~\cite{Fluent} proposes a model composed of a phoneme module, a word module and an intent module that can all be trained or fine-tuned independently. The authors experimented with four settings: with or without pretraining and fine tuning the word module only or all modules together. The two best models (pretrained on ASR and the fine-tuned word module only) are displayed in Tables \ref{tab:fsc}--\ref{tab:mase}. ST-BERT \cite{ST-Bert2020} also uses an MLM objective. However, it is pretrained with cross-modal language modeling on speech and text data in two different ways: using MLM or with conditional language modeling (CLM), where the goal is to predict one modality given the other. Additionally, the model uses large text corpora pretraining and domain adaptation pretraining with the SLU transcripts to further improve the results. However, the use of text transcripts from the downstream task is not compatible with our use case, as we assume that only the speech is available for the SLU task. Since domain adaptation provides an unfair advantage, we do not report the results related to this experiment. The pretrained model corresponds to the model that was pretrained on speech-only data, and the fine-tuned model corresponds to the model pretrained on text with CLM and fine-tuned on SLU. {The performance of these two baselines were reported after training on 1\% and 10\% of the data, which is not the case for the following. Wav2Vec2-Classifier \cite{Seo2022} is an encoder-only model with a fully connected layer as the output. The model is entirely fine-tuned, and the performance when freezing Wav2Vec2 is not reported, although we know from \cite{WangBoumadane2021} that Wav2Vec does not store semantics in its internal representations and that fine tuning is necessary to achieve decent performance. ESPnet-SLU \cite{espnet-slu} uses a pretrained HuBERT as a feature extractor and a transformer decoder for the SLU module. The complete model is fine-tuned on SLU.}

{Table \ref{tab:fsc} shows the results on Fluent Speech Commands when the models are trained on varying amounts of training data. For completeness, we also report the results of the models fine-tuned on the entire training set (Table \ref{tab:fsc100}). In Table \ref{tab:mase}, the challenge splits \cite{MASE} correspond to improved splits, where specific speakers or specific utterances are partitioned.} For Smartlights, no previous splits are available, so we follow the same approach as \cite{MASE} and use a random splitting strategy.

In Table \ref{tab:fsc}, we observe significant improvements compared to \cite{Fluent}, especially when few training examples are used, both before and after fine tuning. {Both \cite{Fluent} and \cite{Seo2022} noted that fine tuning the entire model does not always improve performance because exposition to new data leads to catastrophic forgetting and, in some cases, overfitting.} We make the same observation and find that unfreezing the encoder leads to its degradation. {Indeed, the encoder was pretrained on large amounts of data and is robust to variations in the input features.} When it is fine-tuned with a different objective for a handful of examples, the parameter updates following the new setting often lead to overfitting and loss of generality. {This is also observed for ESPnet-SLU \cite{espnet-slu} in Table \ref{tab:mase}, where we observe a much larger difference between the performance on unknown speakers and unknown utterances than with the other models.} Updating the encoder also deteriorates its ability to generate a good template for the decoder, which leads to weaker performance. Consequently, we decided to keep the encoder frozen but update the decoder during the fine-tuning stage. With this setting, the performance improves slightly, although consistently compared to the model before fine tuning. Our model shows a similar pretraining accuracy as \cite{ST-Bert2020}, but small improvements are observed after fine tuning. {The good performance of the frozen models is encouraging, as it suggests that the pretrained model stores semantic information adequately enough for the SLU module to make use of it. Both approaches using an MLM objective are particularly efficient when trained on 1\% of the training data, which leads us to conclude that bidirectional context embeddings are particularly useful in a low-resource setting.}

{Our model also fares well against \cite{Seo2022} and \cite{espnet-slu}, although our methodology does not include data augmentation (Table \ref{tab:fsc100}). Unfortunately, we cannot compare the quality of the representations produced by the pretrained models, owing to a lack of reported results.}

{In Table \ref{tab:mase}, we observe a 12\% relative improvement compared to \cite{Fluent} on FSC (Challenge) but a deterioration on Smartlights. We speculate that the fact that the MLM objective produces similar embeddings for word pieces that belong to the same semantic scope (such as blue and green or bedroom and bathroom) hurts the performance on this particular dataset.} The few examples in the training set do not suffice to generalize to unknown phrasings for a specific target (i.e., when the same target is expressed differently).

%\begin{table}[H]
    %\caption{Results on Fluent Speech Commands (top) and Smartlights and MASE splits \cite{MASE} (bottom).}
\begin{table}
    \begin{subtable}[t]{.55\linewidth}
        \footnotesize
        \centering
        \begin{tabular}{l|c|c|c|c|c}
            Stage & \% train & MLM & E2E SLU \cite{Fluent} & ST-BERT \cite{ST-Bert2020} \\
            \hline
            \multirow{3}{*}{frozen} & 100 & \textbf{99.5} & 98.8 & 99.4 \\
             & 10 & 99.0 & 98.0 & 99.0 \\
             & 1 & \textbf{91.5} & 82.8 & 89.8 \\
            \hline
            \multirow{3}{*}{finetune} & 100 & \textbf{99.8} & 99.1 & 99.5 \\
             & 10 & \textbf{99.5} & 97.9 & 99.1 \\
             & 1 & 95.7 & - & 95.7 \\
        \end{tabular}
        % \vspace{3mm}
        \caption{Test accuracies on Fluent Speech Commands (original splits). In the first stage, the pretrained model is frozen and the SLU layers are trained. In the second stage, the pretrained model is (partially) finetuned.}
        \label{tab:fsc}
        % \vspace{-3mm}
    \end{subtable}
    \begin{subtable}[t]{.4\linewidth}
        \vspace{-13mm}
        \footnotesize
        \centering
        \begin{tabular}{l|c|c|c|c|c}
            Model & Accuracy \\
            \hline
            MLM (finetune) & \textbf{99.8} \\
            E2E SLU \cite{Fluent} & 99.1 \\
            ST-BERT \cite{ST-Bert2020} & 99.5 \\
            Wav2Vec2-Classifier \cite{Seo2022} & 99.7 \\
            ESPnet-SLU \cite{espnet-slu} & 99.6 \\
        \end{tabular}
        \vspace{4mm}
        \caption{Accuracy on Fluent Speech Commands original test set (100\%) after finetuning.}
        \label{tab:fsc100}
    \end{subtable}
    \begin{subtable}{\linewidth}
        \footnotesize
        \centering
        \begin{tabular}{l|l|c|c|c|c|c}
         &  & \multicolumn{2}{c}{FSC (Ch.)} & Smartlights & \multicolumn{2}{c}{Smartlights (Ch.)} \\
         &  & Spk. & Utt. & Random & Spk. & Utt. \\
        \hline
        \multirow{3}{*}{frozen} & MLM & \textbf{95.2} & \textbf{86.3} & \textbf{84.0} & \textbf{79.5} & \textbf{69.0} \\
            & E2E SLU \cite{Fluent} & 90.9 & 73.4 & 83.2 & 73.2 & 67.4 \\
            & ST-BERT \cite{ST-Bert2020} & - & - & 81.2 & - & - \\
        \hline
        \multirow{4}{*}{finetune} & MLM & \textbf{98.8} & \textbf{88.0} & 85.8 & 81.9 & 74.6 \\
            & E2E SLU \cite{Fluent} & 92.3 & 78.3 & \textbf{88.0} & \textbf{82.6} & \textbf{78.5} \\
            & ST-BERT \cite{ST-Bert2020} & - & - & 84.7 & - & - \\
            & ESPnet-SLU \cite{espnet-slu} & 97.5 & 78.5 & - & - & - \\
        \end{tabular}
        \vspace{3mm}
        \caption{Test accuracies on the Fluent Speech Commands (challenge splits) and Smartlights datasets random and challenge splits. In the first stage, the pretrained model is frozen and the SLU layers are trained. In the second stage, the pretrained model is partially (or totally in \cite{espnet-slu}) finetuned.}
        \label{tab:mase}
    \end{subtable}
    \caption{Results on Fluent Speech Commands (top) and Smartlights and MASE splits \cite{MASE} (bottom)}
    \label{tab:results}
    \vspace{-5mm}
\end{table}

%%%%%%%%%%%%%%%%%%%%%%%%%%%%%%%%%%%%%%%%%%
\section{Analysis}
\label{sec:analysis}

\subsection{Low-Resource Scenario}
\label{sec:analysis:curve}

\begin{figure}[h]
  
    \begin{subfigure}[b]{0.47\linewidth}
        
        \includegraphics[width=\linewidth,keepaspectratio]{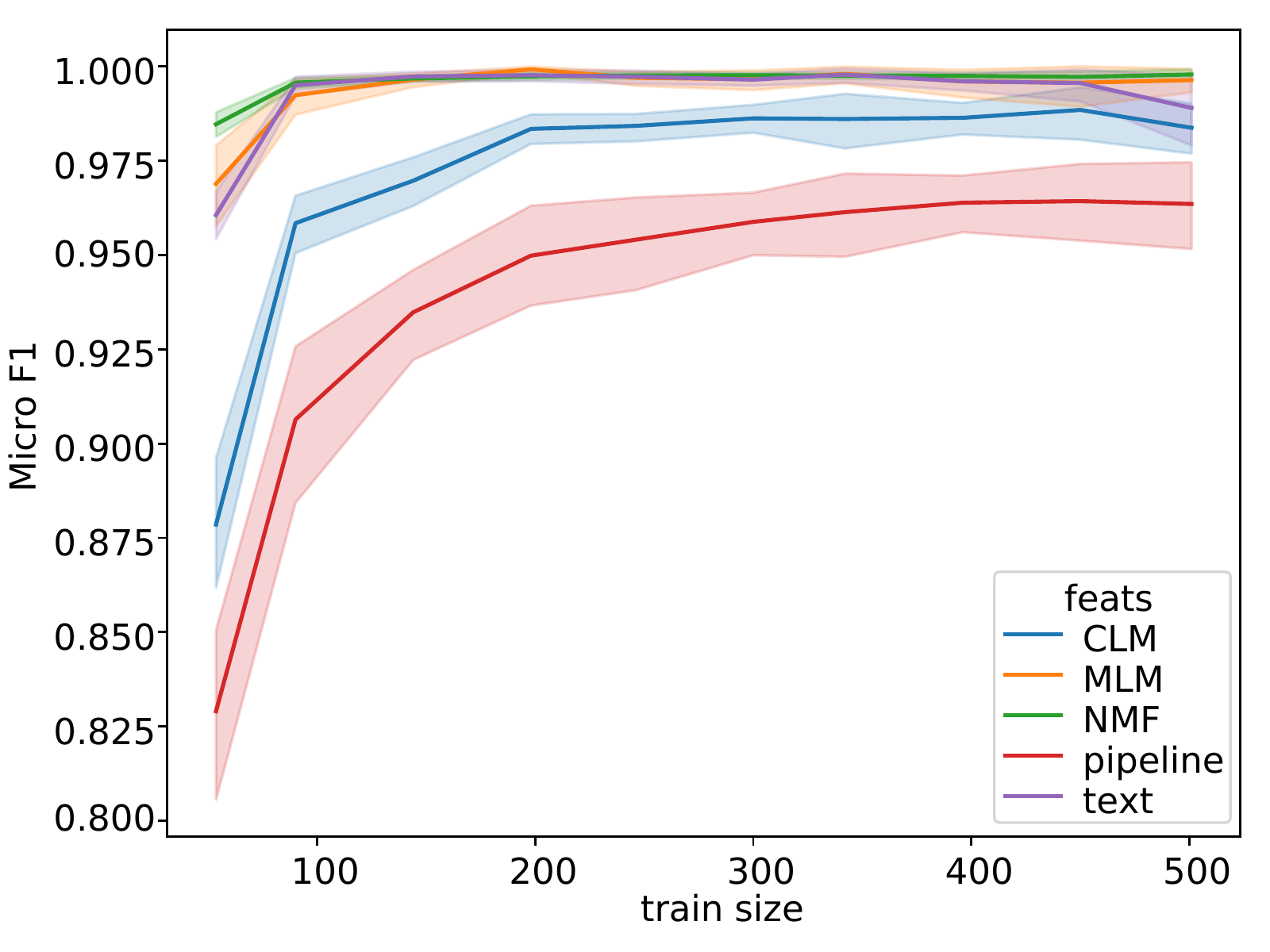}
        \caption{\centering}
        \label{fig:lc:grabo}
    \end{subfigure}
    \begin{subfigure}[b]{0.47\linewidth}       
        \includegraphics[width=\linewidth,keepaspectratio]{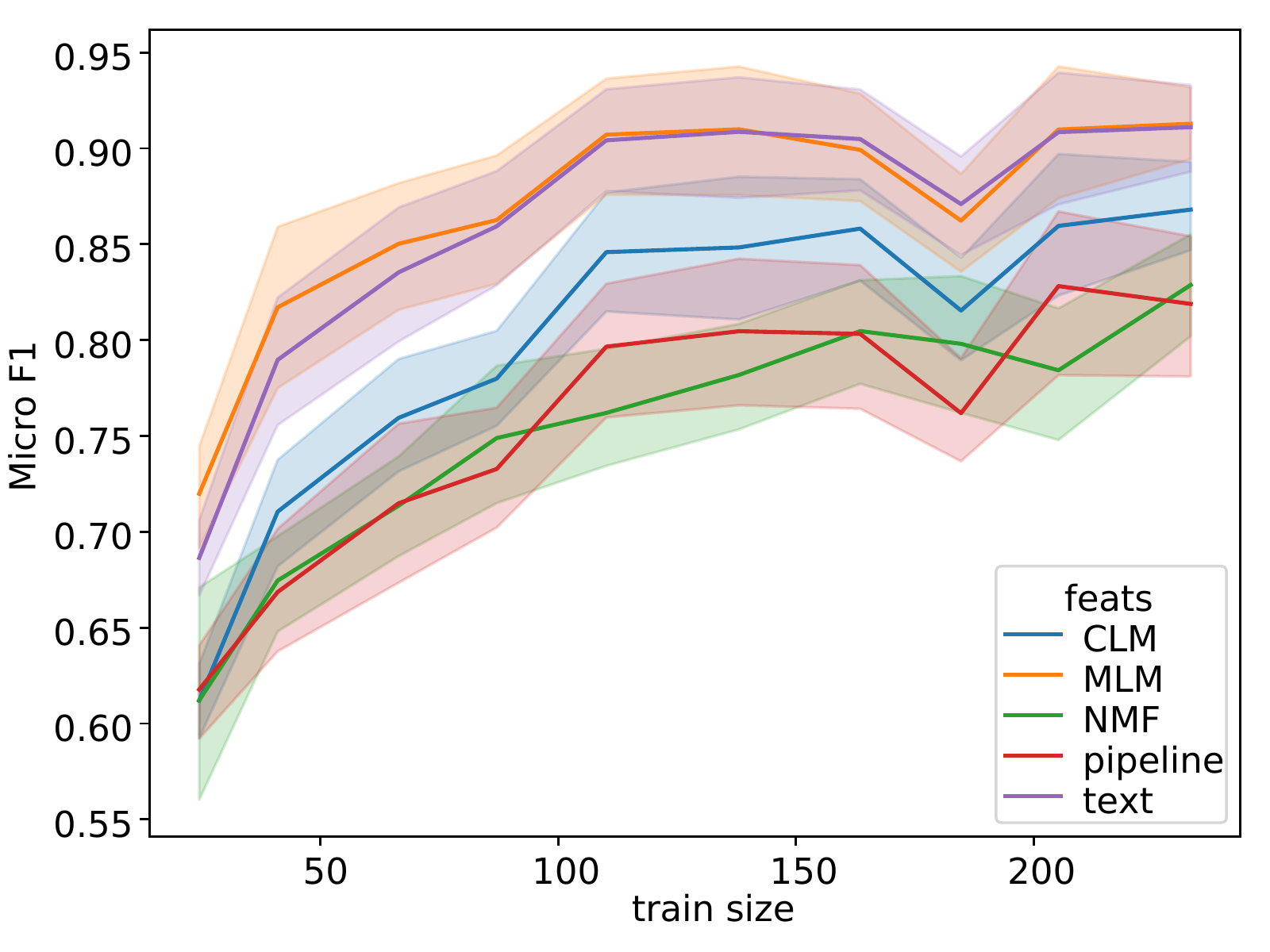}
        \caption{\centering}
        \label{fig:lc:patcor}
    \end{subfigure}
    \caption{
    Comparison of different models trained with increasing train set sizes on (\textbf{a}) Grabo and (\textbf{b}) Patience. Each curve represents the F1 score on the test set as a function of the size of the training set (utterance count). The colored areas represent the 68\% confidence interval. We use the F1 score for comparison with the NMF baseline \cite{Wang2021}.}
    \label{fig:lc}
\end{figure}

To explore low-resource scenarios, we split Grabo and Patience per speaker. For each speaker, we randomly select a fixed number of examples per class for the training set of the SLU model. By gradually increasing the size of the training set, we are able to measure the learning curve. This operation is performed three times per speaker, with different splits each time. We compute the micro-averaged F1 score for each experiment and report the average F1 score with its standard deviation in Figure \ref{fig:lc}.

We compare our representations (MLM in Figure \ref{fig:lc}) with three types of features: NLP features generated by encoding the gold transcriptions with bert-base-dutch-cased \cite{wietsedv2019}, Pipeline features resulting from the encoding of ASR transcripts predicted by ESPnet's hybrid ASR model \cite{hybridctcatt} trained on CGN with bert-base-dutch-cased and CLM features corresponding to the output of the penultimate layer of the ASR model. 
To provide a fair comparison, neither the NLP, ASR nor MLM model encountered training examples from the SLU dataset. In other words, we do not fine tune any of the representation models on the downstream datasets at this stage. Both the ASR and MLM models are trained on CGN, and the NLP model is trained on a collection of five text corpora for a total of about 2.4B tokens.

Considering the amount of training data, it is no surprise to see that the gold transcript encoded with the NLP model performs very well in all data regimes. As expected, the ASR transcripts show the weakest performance. Converting features into discrete symbols forces the model to make decisions, resulting in potential errors from which the model cannot recover. In contrast, the CLM and MLM models use the hidden representations produced by the model, thus avoiding this problem. Nonetheless, we see that the CLM features perform worse than our model. The masked language modeling objective allows the MLM model to look at both right and left contexts, whereas the CLM model only has access to previous predictions. Because each predicted unit is conditioned on the whole sentence, the model is able to derive better representations than by only looking at the past.

We also compare our results with the state of the art on both datasets \cite{Wang2021}. This model combines ASR pretraining and an NMF decoder for intent recognition \cite{Wang2021,Wang2023}. The NMF decoder uses a bag-of-words approach with multihot intent representation, thus ignoring the order of the words in the sentence. Although this model was presented for dysarthric speech, results are available for both the Grabo and Patience datasets \cite{Wang2021}. The main advantage of the NMF decoder is its low computational requirements, which makes it particularly well-suited when very few training data are available, as can be observed in Figure \ref{fig:lc}a. This advantage disappears as the size of the training set increases. However, for the Patience corpus, in which the order of words is important for prediction, our approach achieves better performance compared to the model proposed by \cite{Wang2023}, in which the sequence order is ignored (Figure \ref{fig:lc}b).

\subsection{Representation Content}
\label{sec:analysis:tsne}

\begin{figure}[h]
   
    \begin{subfigure}[b]{0.23\linewidth}
       
        \includegraphics[width=\linewidth,keepaspectratio]{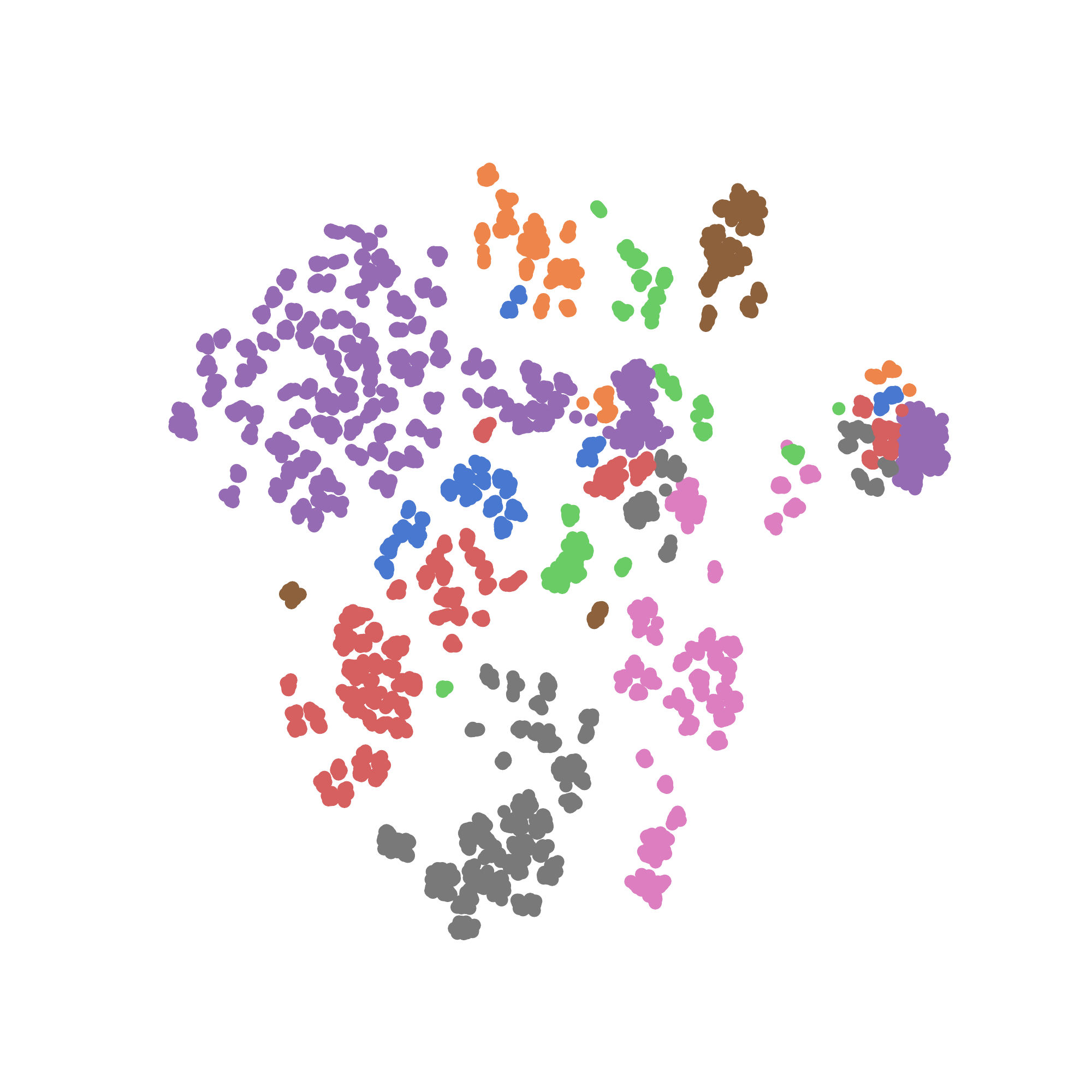}
        \caption{ \centering}
        \label{fig:tsne:enc11}
    \end{subfigure}
    \begin{subfigure}[b]{0.23\linewidth}
       
        \includegraphics[width=\linewidth,keepaspectratio]{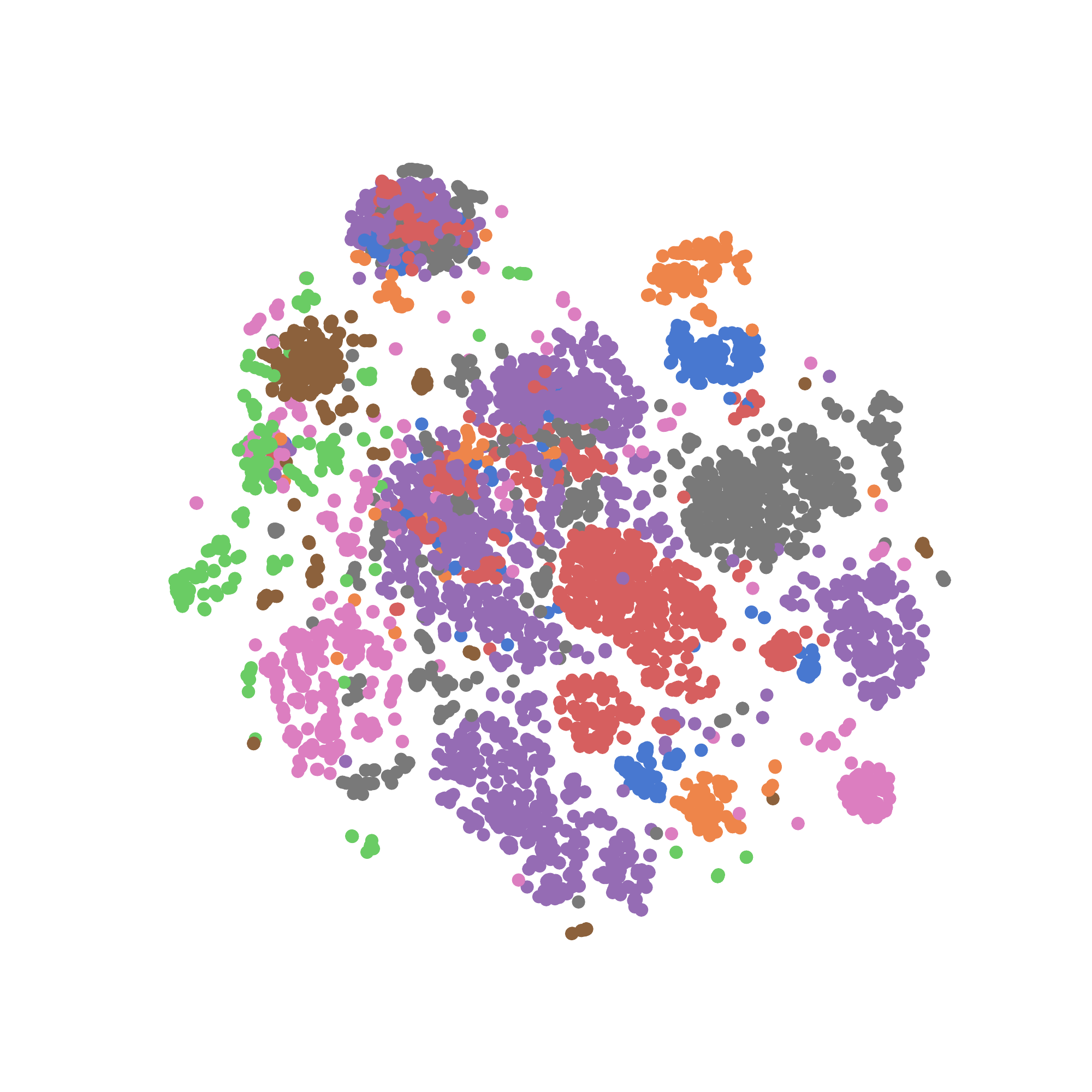}
        \caption{ \centering}
        \label{fig:tsne:dec2}
    \end{subfigure}
    \begin{subfigure}[b]{0.23\linewidth}
       
        \includegraphics[width=\linewidth,keepaspectratio]{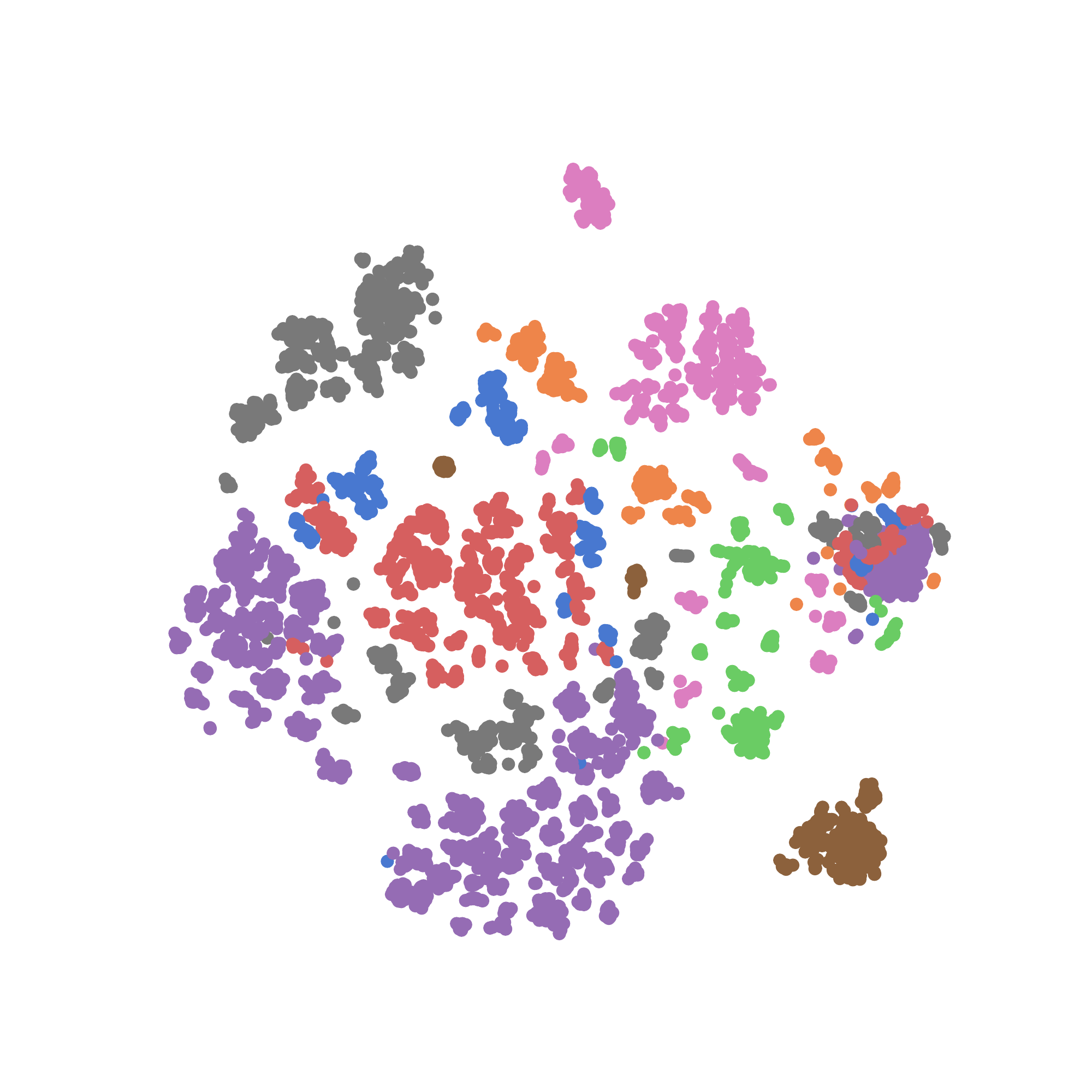}
        \caption{ \centering}
        \label{fig:tsne:dec5}
    \end{subfigure}
    \begin{subfigure}[b]{0.23\linewidth}
       
        \includegraphics[width=\linewidth,keepaspectratio]{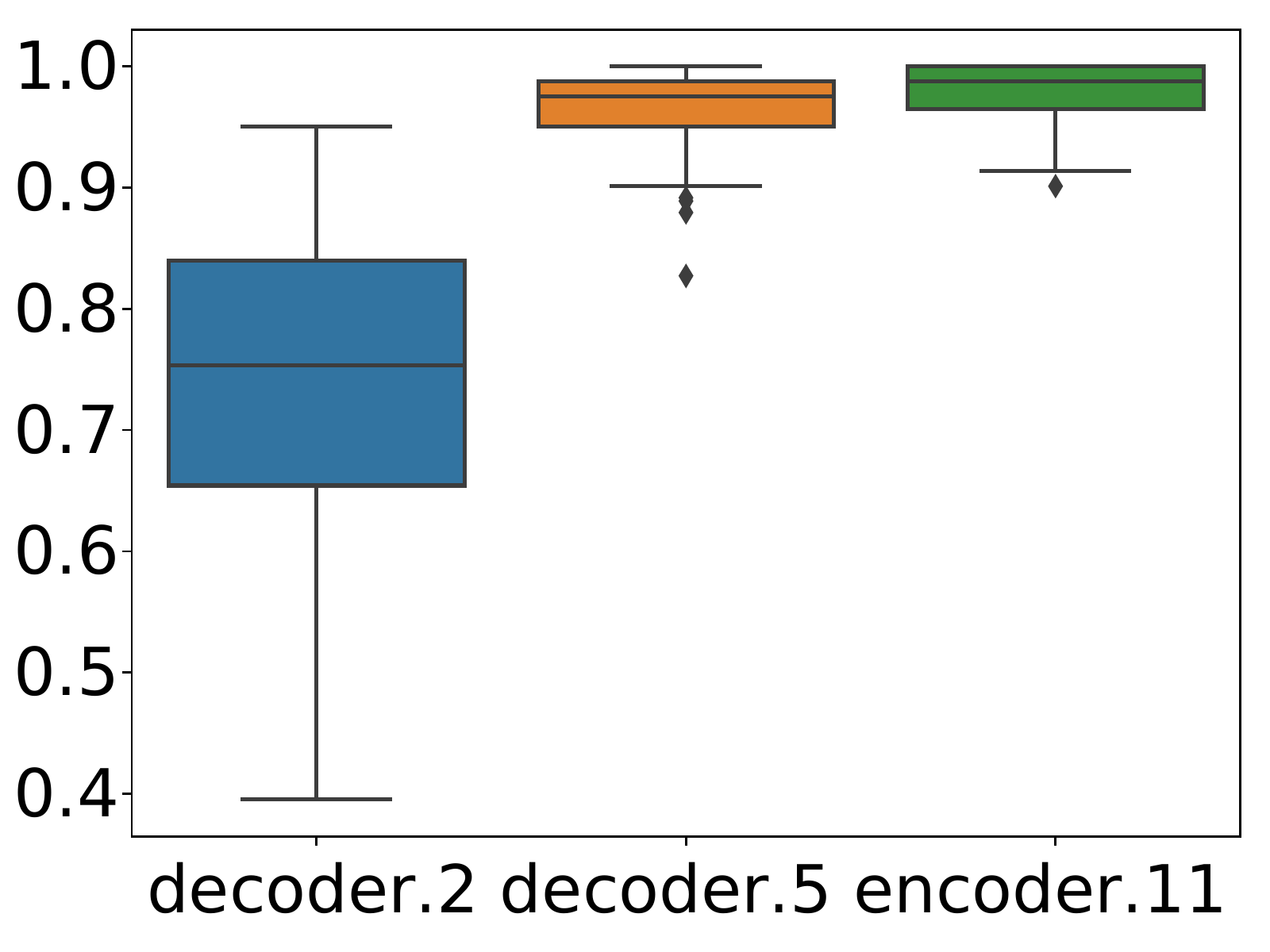}
        \caption{ \centering}
        \label{fig:tsne:bp}
    \end{subfigure}
    \caption{t-SNE representations in different layers. Intents: approach (blue), grab (orange), lift (green), move\_abs (red), move\_rel (purple), pointer (brown), turn\_abs (pink) and turn\_rel (grey). The right-most graph shows the accuracy of the models with the representations of different layers as input. Best viewed in color. (\textbf{a}) Encoder.11; (\textbf{b}) decoder.2; (\textbf{c}) decoder.5; (\textbf{d}) SLU accuracy.}
    \label{fig:tsne}
\end{figure}

To better understand the characteristics of the representations, we average the sequences to a unique representation per utterance and visualize them in two dimensions using the t-SNE algorithm \cite{tsne}. For this experiment, we focus on Grabo and, in particular, on the eight actions that the robot can perform. We compare the representations produced by different layers, namely the last encoder layer, the third decoder layer and the last decoder layer  (encoder.11, decoder.2 and decoder.5 in Figure \ref{fig:tsne}, respectively). Both encoder.11 and decoder.5 form meaningful clusters, while decoder.2 is ill-defined. This is also observed in the classification accuracy of those representations on SLU, where we observe a clear difference between decoder.2 and the other features. We expected that decoder.5 would generate better features than encoder.11, although they seem to lead to similar performance, with a small advantage for the encoder's output. We conjecture that the CTC component helps to define the acoustic units at the encoder's output, which translates into well-defined representation sequences, albeit much longer ones.

We also observe that approach and move\_abs end up in the same region, which means that a classifier might often confuse them. This makes sense from a language perspective, and the fact that this is more the case in decoder.5 than in encoder.11 leads us to conclude that the implicit language model from the decoder generates similar embeddings for these two commands, although they do not sound similar.

\subsection{Interpreting Model Predictions}
\label{sec:analysis:cla}

\begin{figure}[h]

    \includegraphics[width=.95\linewidth]{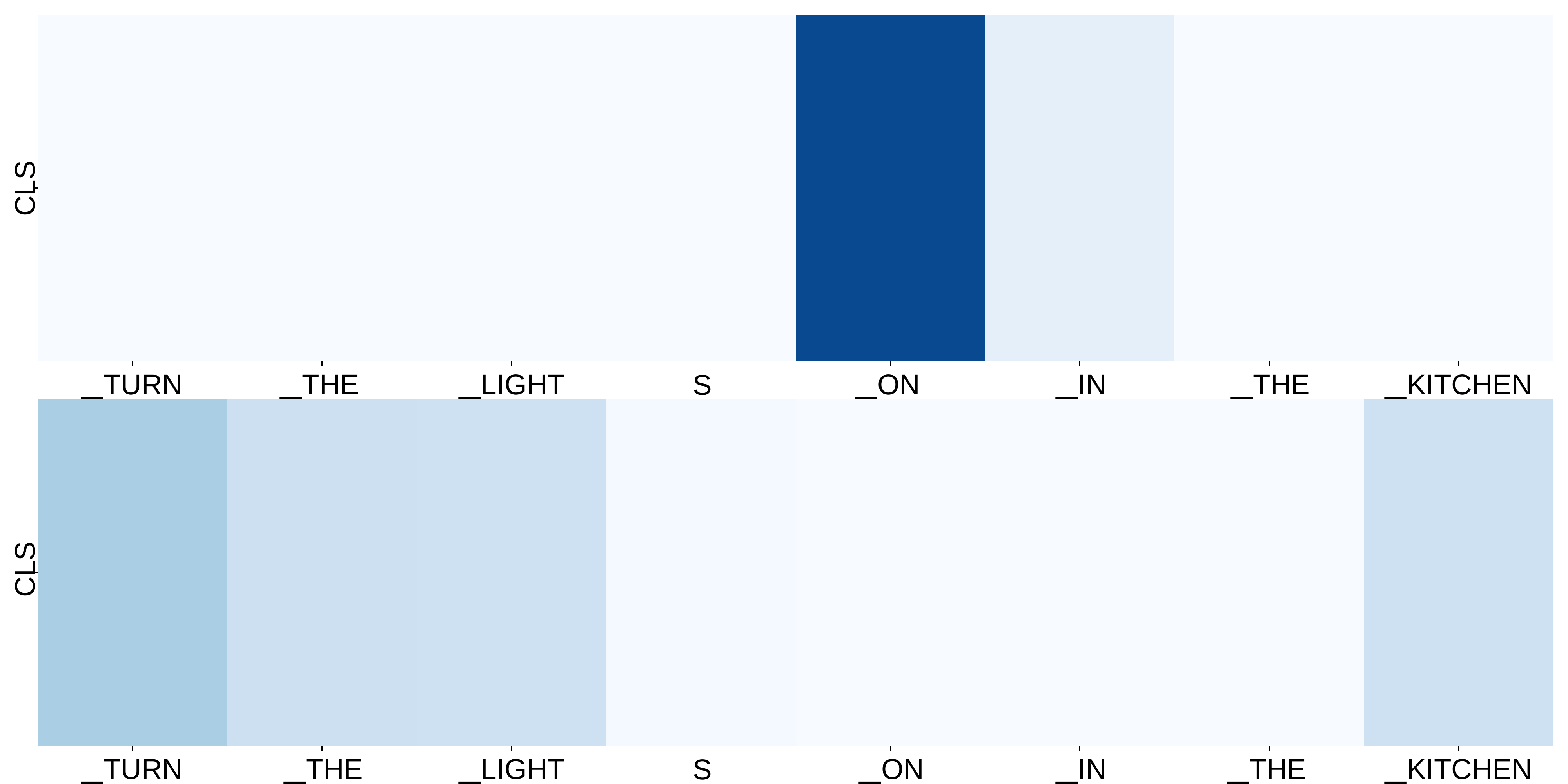}
    \caption{Class attention weight visualization of the two layers of the downstream class attention model. Although the model receives sequences of embeddings, we label the graph with the corresponding tokens for demonstration purposes. The two layers focus on different positions to predict the intent and arguments.}
    \label{fig:cla}
\end{figure}

{Finally, we want to explore an aspect that is often overlooked in deep learning, namely elucidating the factors in the input features that drive the model's predictions. A key aspect of our class attention layer is its ability to establish a connection between the final output prediction and the input sequence through the attention weights. These weights can be interpreted as the relevance of each input unit in determining the correct label for the utterance. As shown in Figure \ref{fig:cla}, a limited subset of tokens (e.g., ``turn'', ``the'', ``light'', ``on'' and ``kitchen'') substantially influences the prediction process. In instances in which the model's predictions may deviate in terms of accuracy, these attention weights serve as valuable starting points for in-depth investigations into the underlying causes.}

%%%%%%%%%%%%%%%%%%%%%%%%%%%%%%%%%%%%%%%%%%
\section{Conclusions}
{Despite the significant advancements in speech processing, there remains a crucial gap in the field concerning the development of speech representation models that capture semantics akin to the capabilities exhibited by language models. In this paper, we introduce a novel bidirectional representation model pretrained on an ASR task, which demonstrates remarkable effectiveness in transferring learned features to the domain of intent recognition, all without necessitating additional fine tuning. Our pretraining strategy incorporates CTC, together with an MLM objective, yielding speech features endowed with bidirectional contextual awareness.}

{Throughout our research, we examined the representations in various layers of the model, particularly with regard to their performance when applied to previously unseen datasets. Our findings show the versatility of these representations in training an effective SLU model. Notably, our representations either match or surpass the performance of state-of-the-art models on the SLU task, especially prior to the fine-tuning phase. We have also introduced a novel approach leveraging class attention mechanisms to summarize sequences of representations. This approach not only proves highly efficient, owing to its transformer-like architecture and minimal parameter requirements, but also offers insights into the model's decision-making process, shedding light on the input patterns that significantly influence predictions. Furthermore, our combination of class attention and ASR pretraining exhibits substantial gains in terms of data efficiency, as evidenced in a low-resource scenario in which we intentionally constrained the number of training examples available for the SLU task. In an era of increasingly massive and data-hungry models, our research underscores the importance of pursuing resource-friendly and efficient solutions.}

{However, it is important to acknowledge a limitation of our presented model, namely its challenges when confronted with SLU datasets that differentiate between terms seldom encountered or entirely absent in the pretraining data. In such cases in which two words are largely interchangeable except for a few instances in the pretraining dataset, our model's representations may become overly similar, hindering the ability to discriminate between these distinct classes without fine tuning. Addressing this limitation will be a focal point of our future research endeavors.}

\section*{Declarations}

The authors have no conflicts of interest to declare that are relevant to the content of this article.

\section*{Acknowledgments}
This research received funding from the Flemish Government under the ``Onderzoeksprogramma Artificiele Intelligentie (AI) Vlaanderen'' programme.

%Bibliography
\bibliographystyle{unsrt}  
\bibliography{references}  

\begin{thebibliography}{10}

\bibitem{Vaswani2017}
Ashish Vaswani, Noam Shazeer, Niki Parmar, Jakob Uszkoreit, Llion Jones, Aidan~N Gomez, Lukasz Kaiser, and Illia Polosukhin.
\newblock Attention is all you need.
\newblock In {\em Advances in Neural Information Processing Systems}. Curran Associates, 2017.

\bibitem{jawahar-etal-2019-bert}
Ganesh Jawahar, Beno{\^\i}t Sagot, and Djam{\'e} Seddah.
\newblock What does {BERT} learn about the structure of language?
\newblock In {\em Proceedings of the 57th Annual Meeting of the Association for Computational Linguistics}. Association for Computational Linguistics, 2019.

\bibitem{Devlin2019}
Jacob Devlin, Ming-Wei Chang, Kenton Lee, and Kristina Toutanova.
\newblock {BERT}: Pre-training of deep bidirectional transformers for language understanding.
\newblock In {\em Proceedings of the 2019 Conference of the North {A}merican Chapter of the Association for Computational Linguistics: Human Language Technologies, Volume 1 (Long and Short Papers)}. Association for Computing Machinery, 2019.

\bibitem{BECTRA}
Yosuke Higuchi, Tetsuji Ogawa, Tetsunori Kobayashi, and Shinji Watanabe.
\newblock Bectra: Transducer-based end-to-end asr with bert-enhanced encoder.
\newblock In {\em IEEE International Conference on Acoustics, Speech and Signal Processing (ICASSP)}. IEEE Signal Processing Society, 2023.

\bibitem{Karita2019Compare}
Shigeki Karita, Xiaofei Wang, Shinji Watanabe, Takenori Yoshimura, Wangyou Zhang, Nanxin Chen, Tomoki Hayashi, Takaaki Hori, Hirofumi Inaguma, Ziyan Jiang, Masao Someki, Nelson Yalta, and Ryuichi Yamamoto.
\newblock A comparative study on transformer vs rnn in speech applications.
\newblock In {\em Automatic Speech Recognition and Understanding Workshop (ASRU)}. IEEE Signal Processing Society, 2019.

\bibitem{Dong2018}
Linhao Dong, Shuang Xu, and Bo~Xu.
\newblock {Speech-Transformer: A No-Recurrence Sequence-to-Sequence Model for Speech Recognition}.
\newblock In {\em IEEE International Conference on Acoustics, Speech and Signal Processing (ICASSP)}. IEEE Signal Processing Society, 2018.

\bibitem{Mohamed2020}
Abdelrahman Mohamed, Dmytro Okhonko, and Luke Zettlemoyer.
\newblock Transformers with convolutional context for {ASR}.
\newblock {\em Computing Research Repository}, 2019.

\bibitem{wav2vec2}
Alexei Baevski, Henry Zhou, Abdelrahman Mohamed, and Michael Auli.
\newblock Wav2vec 2.0: A framework for self-supervised learning of speech representations.
\newblock In {\em Proceedings of the 34th International Conference on Neural Information Processing Systems}. Curran Associates, 2020.

\bibitem{hubert}
Wei-Ning Hsu, Benjamin Bolte, Yao-Hung Tsai, Kushal Lakhotia, Ruslan Salakhutdinov, and Abdelrahman Mohamed.
\newblock Hubert: Self-supervised speech representation learning by masked prediction of hidden units.
\newblock {\em IEEE/ACM Transactions on Audio, Speech, and Language Processing}, 2021.

\bibitem{WangBoumadane2021}
Yingzhi Wang, Abdelmoumene Boumadane, and Abdelwahab Heba.
\newblock A fine-tuned wav2vec 2.0/hubert benchmark for speech emotion recognition, speaker verification and spoken language understanding.
\newblock {\em Computing Research Repository}, 2021.

\bibitem{espnet-slu}
Shinji Watanabe, Takaaki Hori, Shigeki Karita, Tomoki Hayashi, Jiro Nishitoba, et~al.
\newblock Espnet: End-to-end speech processing toolkit.
\newblock In {\em Proc. Interspeech}. International Speech Communication Association, 2018.

\bibitem{hybridctcatt}
Shinji Watanabe, Takaaki Hori, Suyoun Kim, John~R. Hershey, and Tomoki Hayashi.
\newblock Hybrid ctc/attention architecture for end-to-end speech recognition.
\newblock {\em IEEE Journal of Selected Topics in Signal Processing}, 2017.

\bibitem{Graves2006}
Alex Graves, Santiago Fern\'{a}ndez, Faustino Gomez, and J\"{u}rgen Schmidhuber.
\newblock Connectionist temporal classification: Labelling unsegmented sequence data with recurrent neural networks.
\newblock In {\em Proceedings of the 23rd International Conference on Machine Learning}. Association for Computing Machinery, 2006.

\bibitem{gao2022}
Heting Gao, Junrui Ni, Kaizhi Qian, Yang Zhang, Shiyu Chang, and Mark Hasegawa-Johnson.
\newblock {WavPrompt: Towards Few-Shot Spoken Language Understanding with Frozen Language Models}.
\newblock In {\em Proc. Interspeech}, 2022.

\bibitem{MaskCTC2020}
Yosuke Higuchi, Shinji Watanabe, Nanxin Chen, Tetsuji Ogawa, and Tetsunori Kobayashi.
\newblock Mask {CTC:} non-autoregressive end-to-end {ASR} with {CTC} and mask predict.
\newblock In {\em Proc. Interspeech 2020}. International Speech Communication Association, 2020.

\bibitem{ctcdecode}
Thomas Zenkel, Ramon Sanabria, Florian Metze, Jan Niehues, Matthias Sperber, Sebastian Stüker, and Alex Waibel.
\newblock Comparison of decoding strategies for ctc acoustic models.
\newblock In {\em Proc. Interspeech 2017}. International Speech Communication Association, 2017.

\bibitem{ELMo}
Matthew~E. Peters, Mark Neumann, Mohit Iyyer, Matt Gardner, Christopher Clark, Kenton Lee, and Luke Zettlemoyer.
\newblock Deep contextualized word representations.
\newblock In {\em Proceedings of the 2018 Conference of the North {A}merican Chapter of the Association for Computational Linguistics}. Association for Computational Linguistics, 2018.

\bibitem{MaskPredict}
Marjan Ghazvininejad, Omer Levy, Yinhan Liu, and Luke Zettlemoyer.
\newblock Mask-predict: Parallel decoding of conditional masked language models.
\newblock In {\em Proceedings of the 2019 Conference on Empirical Methods in Natural Language Processing and the 9th International Joint Conference on Natural Language Processing (EMNLP-IJCNLP)}. Association for Computational Linguistics, 2019.

\bibitem{ClassAtt}
Hugo Touvron, Matthieu Cord, Alexandre Sablayrolles, Gabriel Synnaeve, and Herv{\'{e}} J{\'{e}}gou.
\newblock Going deeper with image transformers.
\newblock In {\em 2021 IEEE/CVF International Conference on Computer Vision (ICCV)}. IEEE Computer Society, 2021.

\bibitem{LayerNorm}
Ruibin Xiong, Yunchang Yang, Di~He, Kai Zheng, Shuxin Zheng, Chen Xing, Huishuai Zhang, Yanyan Lan, Liwei Wang, and Tie-Yan Liu.
\newblock On layer normalization in the transformer architecture.
\newblock In {\em Proceedings of the 37th International Conference on Machine Learning}. JMLR.org, 2020.

\bibitem{CGN}
N.H.J. Oostdijk.
\newblock Het corpus gesproken nederlands.
\newblock {\em Nederlandse Taalkunde}, 2000.

\bibitem{Librispeech}
Vassil Panayotov, Guoguo Chen, Daniel Povey, and Sanjeev Khudanpur.
\newblock Librispeech: An asr corpus based on public domain audio books.
\newblock In {\em IEEE International Conference on Acoustics, Speech and Signal Processing (ICASSP)}. IEEE Signal Processing Society, 2015.

\bibitem{Renkens2018}
Vincent Renkens and Hugo Van~hamme.
\newblock Capsule networks for low resource spoken language understanding.
\newblock In {\em Proc. Interspeech}. International Speech Communication Association, 2018.

\bibitem{Patcor}
Netsanet Tessema, Bart Ons, Janneke van~de Loo, Jort~F. Gemmeke, Guy De~Pauw, Walter Daelemans, and Hugo Van~hamme.
\newblock Metadata for corpora patcor and domotica-2.
\newblock 2013.

\bibitem{Fluent}
Loren Lugosch, Mirco Ravanelli, Patrick Ignoto, Vikrant~Singh Tomar, and Yoshua Bengio.
\newblock {Speech Model Pre-Training for End-to-End Spoken Language Understanding}.
\newblock In {\em Proc. Interspeech}. International Speech Communication Association, 2019.

\bibitem{MASE}
Siddhant Arora, Alissa Ostapenko, Vijay Viswanathan, Siddharth Dalmia, Florian Metze, Shinji Watanabe, and Alan~W Black.
\newblock Rethinking end-to-end evaluation of decomposable tasks: A case study on spoken language understanding.
\newblock In {\em Proc. Interspeech}. International Speech Communication Association, 2021.

\bibitem{Smartlights}
Alaa Saade, Joseph Dureau, David Leroy, Francesco Caltagirone, Alice Coucke, Adrien Ball, Clément Doumouro, Thibaut Lavril, Alexandre Caulier, Théodore Bluche, Thibault Gisselbrecht, and Maël Primet.
\newblock Spoken language understanding on the edge.
\newblock In {\em Fifth Workshop on Energy Efficient Machine Learning and Cognitive Computing - NeurIPS Edition (EMC2-NIPS)}, 2019.

\bibitem{Kingma2017}
Diederik~P. Kingma and Jimmy Ba.
\newblock Adam: {A} method for stochastic optimization.
\newblock In {\em 3rd International Conference on Learning Representations, {ICLR}}, 2015.

\bibitem{tsne}
Laurens van~der Maaten and Geoffrey~E. Hinton.
\newblock Visualizing high-dimensional data using t-sne.
\newblock {\em Journal of Machine Learning Research}, 2008.

\bibitem{ST-Bert2020}
Minjeong Kim, Gyuwan Kim, Sang-Woo Lee, and Jung-Woo Ha.
\newblock {ST-BERT}: Cross-modal language model pre-training for end-to-end spoken language understanding.
\newblock In {\em IEEE International Conference on Acoustics, Speech and Signal Processing (ICASSP)}. IEEE Signal Processing Society, 2021.

\bibitem{Seo2022}
Seunghyun Seo, Donghyun Kwak, and Bowon Lee.
\newblock Integration of pre-trained networks with continuous token interface for end-to-end spoken language understanding.
\newblock In {\em IEEE International Conference on Acoustics, Speech and Signal Processing (ICASSP)}. IEEE Signal Processing Society, 2022.

\bibitem{Wang2021}
Pu~Wang and Hugo Van~hamme.
\newblock Pre-training for low resource speech-to-intent applications.
\newblock 2021.

\bibitem{wietsedv2019}
Wietse de~Vries, Andreas van Cranenburgh, Arianna Bisazza, Tommaso Caselli, Gertjan van Noord, and Malvina Nissim.
\newblock Bertje: {A} dutch {BERT} model.
\newblock {\em Computing Research Repository}, 2019.

\bibitem{Wang2023}
Pu~Wang and Hugo Van~hamme.
\newblock Benefits of pre-trained mono- and cross-lingual speech representations for spoken language understanding of dutch dysarthric speech.
\newblock {\em EURASIP Journal on Audio, Speech, and Music Processing}, 2023.

\end{thebibliography}

\end{document}